\def\year{2020}\relax
\documentclass[letterpaper]{article} 
\usepackage{aaai20}  
\usepackage{times}  
\usepackage{helvet} 
\usepackage{courier}  
\usepackage[hyphens]{url}  
\usepackage{graphicx} 
\usepackage{graphicx}  
\usepackage{lipsum} 
\usepackage{mathtools}
\usepackage{amsmath} 
\usepackage{amssymb}  
\usepackage{algorithm}

\usepackage{bigdelim}
\usepackage{booktabs}
\usepackage{graphicx}
\usepackage{gensymb}
\usepackage{multirow}
\usepackage{todonotes}
\usepackage{url}
\usepackage{breqn}
\usepackage{subcaption}
\usepackage{amsfonts}
\usepackage{xcolor}
\usepackage{array,multirow}
\usepackage{soul}
\usepackage[normalem]{ulem}
\newcommand{\stkout}[1]{\ifmmode\text{\sout{\ensuremath{#1}}}\else\sout{#1}\fi}

\newcommand{\Dgreen}{\textcolor{black}}

\usepackage{xcolor}
\usepackage{xparse}

\NewDocumentCommand{\colornucleus}{omme{_^}}{%
  \begingroup\colorlet{currcolor}{.}%
  \IfValueTF{#1}
   {\textcolor[#1]{#2}}
   {\textcolor{#2}}
    {%
     #3
     \IfValueT{#4}{_{\textcolor{currcolor}{#4}}}
     \IfValueT{#5}{^{\textcolor{currcolor}{#5}}}
    }%
  \endgroup
}

\graphicspath{{./figures/}}

\title{Human-Supervised Semi-Autonomous Mobile Manipulators for Safely and Efficiently Executing Machine Tending Tasks }

\author{Sarah  Al-Hussaini\textsuperscript{\rm 1},
Shantanu Thakar\textsuperscript{\rm 1}, 
Hyojeong Kim\textsuperscript{\rm 1},
Pradeep Rajendran\textsuperscript{\rm 1},\\
\bf \Large{Brual C. Shah\textsuperscript{\rm 1}, Alec Kanyuck\textsuperscript{\rm 1}, 
Jeremy A. Marvel\textsuperscript{\rm 2}, 
Satyandra K. Gupta\textsuperscript{\rm 1}}\\

\textsuperscript{\rm 1}Viterbi School of Engineering,
    University of Southern California, CA, USA\\
\{salhussa, sthakar, hyojeonk, pradeepr, brualsha, kanyuck,  guptask\}@usc.edu\\

\textsuperscript{\rm 2}National Institute of Standards and Technology, Gaithersburg, MD, USA
jeremy.marvel@nist.gov 
}

\begin{document}

\maketitle

\begin{abstract}
Mobile manipulators can be used for machine tending and material handling tasks in small volume manufacturing applications. These applications usually have semi-structured work environment. The use of a fully autonomous mobile manipulator for such applications can be risky, as an inaccurate model of the workspace may result in damage to expensive equipment. On the other hand, the use of a fully teleoperated mobile manipulator may require a significant amount of operator time. In this paper, a semi-autonomous mobile manipulator is developed for safely and efficiently carrying out machine tending tasks under human supervision. The robot is capable of generating motion plans from the high-level task description and presenting simulation results to the human for approval. The human operator can authorize the robot to execute the automatically generated plan or provide additional input to the planner to refine the plan. If the level of uncertainty in some parts of the workspace model is high, then the human can decide to perform teleoperation to safely execute the task. Our preliminary user trials show that non-expert operators can quickly learn to use the system and perform machine tending tasks.
\end{abstract}

\section{Introduction} \label{sec:intro}
In large volume manufacturing, material handling is highly automated using conveyor belts, automated guided vehicles (AGVs), and large industrial robotic arms. This equipment can be expensive and largely inflexible in terms of handling process uncertainty, and thus have limited utility in small-volume, batch manufacturing. For such environments, the majority of material handling is largely handled by humans~\cite{rey2019}. 
Full automation may be infeasible given the shortness of the production runs and the high frequency of manufactured part turnover. 


Robotic manipulators may provide such flexibility through the use of tool changers and multi-purpose tooling.  With the increased focus on computer-numerical-controlled (CNC) machine tools and additive manufacturing technologies, machine tending operations would require dedicated robots for inserting, manipulating, and removing parts.  
However, having a fixed manipulator for every machine is expensive, and not economically viable since the manipulator would be idle for a significant portion of the manufacturing process. As such, a more flexible and versatile system would be needed for flexible machine tending in small-volume applications.

One such system is a mobile manipulator, which is a robotic manipulator integrated onto a mobile platform, frequently with integrated vision and tooling systems. The mobility enhances the operational capabilities and the efficiency of the manipulator, and the ability to physically interact with the environment expands the versatility of the mobile platform. More importantly, the juxtaposition of mobility and manipulation enables cost-efficient operations, as a single robotic system can now tend to multiple machine tools and minimize idle time. 

\begin{figure}[t]
  \centering
  \includegraphics[width=0.7\columnwidth]{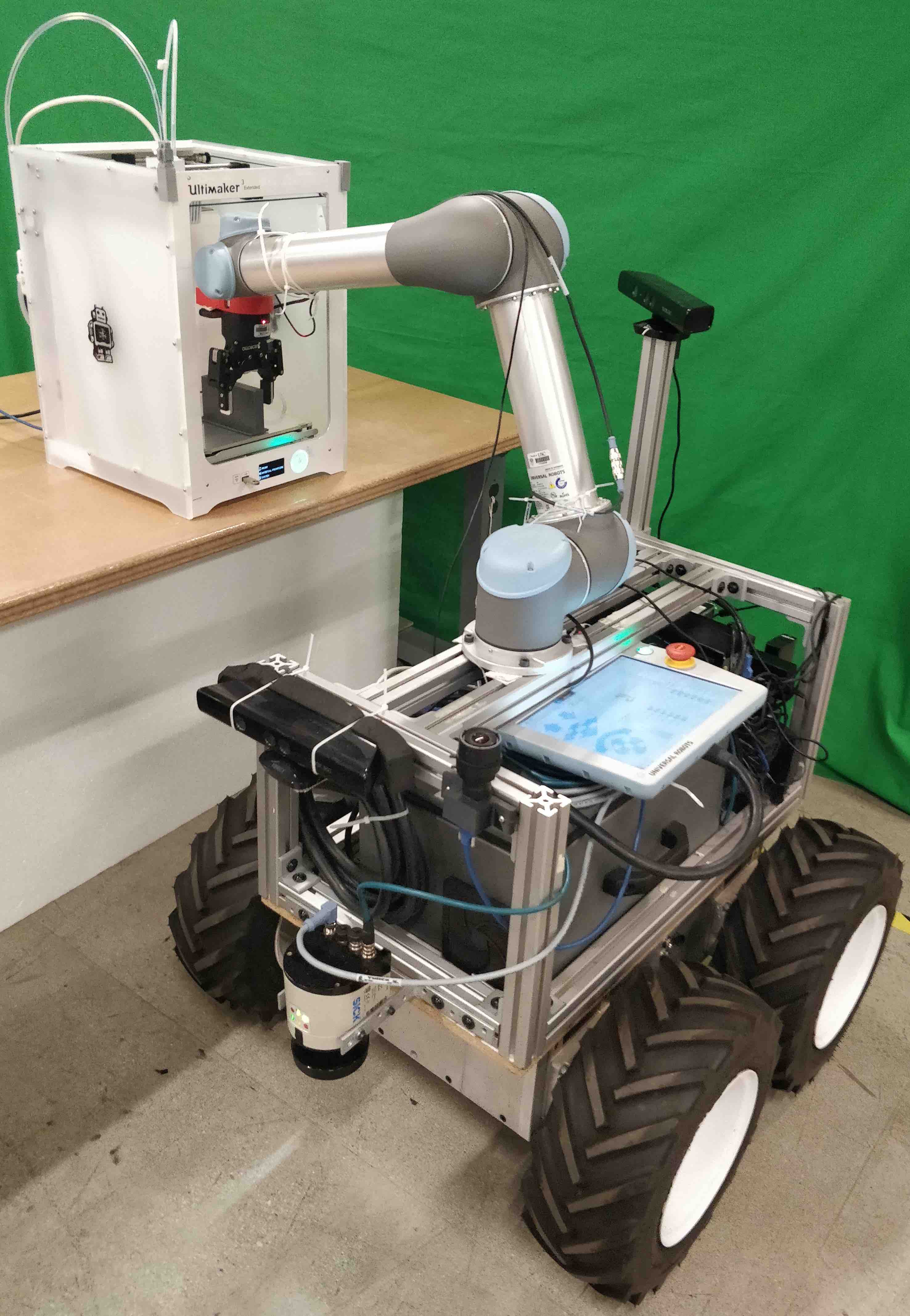}
  \caption{The ADAMMS (Agile Dexterous Autonomous Mobile Manipulation System, ~\cite{annem2019msec}) mobile manipulator system doing a machine tending task.}
  \label{fig:system_hardware}
\end{figure}

Many different types of mobile manipulators have been developed ~\cite{hamner2010autonomous,thakar2018CASE,katz2006umass,domel2017toward,srinivasa2012herb,annem2019msec,hernandez2016teleoperation,thakar2019case}. The operational capabilities of these systems range from being exclusively teleoperated, to fully autonomous. Pure teleoperation implies that the operator has control over both individual joints of the manipulator, and the forward and backward velocities of the mobile base. Such operation gives the operator complete control over the entire robot in an environment with expensive machines, which is highly desirable. However, teleoperating such a system can be tedious as the motions of a manipulator on the joint level are non-intuitive for humans. Further, Cartesian space teleoperation of the manipulator and its gripper may result in unrealistic joint configurations, in terms of collisions or singularities. 

Fully autonomous operations of mobile manipulators may also be applied to machine tending tasks~\cite{thakar2019accounting,thakar2020tase,rajendran2020jcise,Rajendran2019b}. Here, the operator provides only the task goals, and the robot plans the trajectory for the mobile base and the manipulator as well as the grasping for the object using artificial intelligence (AI) technologies~\cite{thakar2020iros,Rajendran2019a,kumbla2017simulation,kumbla2018handling,kabir2019icra,ICRA_2020_Kabir,colombo2019case}. However, such operations can be risky in the presence of sensing uncertainty. It may not be feasible to build a high fidelity model of the environment in semi-structured environments. An inaccurate model of the environment may result in collision and damage to expensive equipment resulting. Hence, completely autonomous operations may not be desirable. 

Teleoperation mode provides safety in terms of collision with expensive equipment, and the autonomous model is efficient as the operator does not make low-level decisions for the system. This paper introduces a human-supervised, semi-autonomous, mobile manipulation system for machine tending operations. A hybrid operation mode is developed in which teleoperation and autonomous motions are combined. For such a system, a remote operator only provides a set of high-level instructions in each individual task. The robot autonomously plans motions for executing these tasks and shows a simulation of plan execution to the operator, who can then chose to execute or discard the motion plans. Moreover, the remote operator can also monitor the motions being executed and stop if a collision is likely to occur. Furthermore, if the autonomous operation is infeasible, the operator can take complete control of the robot and still complete the task in teleoperation mode. As automation in manufacturing applications continues to increase, human operator support of the equipment is expected to become increasingly remote due to spatial, logistical, and safety constraints.  For this reason, the intended purpose is to develop a system to enable a single operator to support multiple platforms distributed throughout a facility through a semi-autonomous operation.

This paper presents system requirements for a mobile manipulator robot to perform machine tending tasks in small volume manufacturing applications based on exploratory user trials. The system architecture and platform design to meet the identified requirements are also discussed. Evaluation of new features of the system is presented through preliminary experimentation.

\section{Related Work} \label{sec:related}

There have been several works on human-assisted operation and teleoperation of robots for the execution of various kind of tasks~\cite{rosen2018testing,orekhov2016analysis,malysz2013task,kim2014experimental,ferraguti2015energy,saeidi2016trust,al2019alert,IROS_2020_Al-Hussaini,burrell2018towards,Isop2019}. Complete teleoperation (i.e., joint level control using haptics) is typically found in surgical robots~\cite{koh2018efficacy} where total control of the robot by human is of prime importance. Complete autonomous operations are typically found in large warehouses where there is order maintained~\cite{d2012guest}. Such operations can assist humans by reducing their workload from mundane tasks.  Teleoperation of mobile manipulators has been studied in~\cite{garcia2019object}, where obstacle avoidance and manipulator dexterity are taken into consideration by exploiting the system redundancy to make the human operation more intuitive and safe.

Controlling a high degree of freedom systems like manipulators using teleoperation is challenging. 
The focus of such operation is typically to manipulate objects or move the end-effector in certain desired ways, which can be achieved via multiple joint configurations. However, moving each joint manually to achieve a certain end-effector position and orientation is non-intuitive. A graphical user interface (GUI) has been presented for positioning, orienting, and actuating a gripper on a manipulator to interact with surrounding objects in~\cite{Kent2017}. 

Haptic feedback can be used for intuitive teleoperation of mobile robots and mobile manipulators ~\cite{Masone2018,Saeidi2017,wrock2017automatic}. This enhances human capabilities since human intelligence can be easily transferred to robots. Haptic based feedback and control can be integrated with semi-autonomous teleoperation as well like in~\cite{YLKent2017}. Here, the humans provide instructions for the motion of the two robots using a haptic feedback control device.  Whereas, autonomy is used to maintain the nonholonomic constraints of two mobile robots cooperating in a transportation task. This is necessary as maintaining nonholonomic constraints during such complex tasks is challenging for humans. Both virtual reality and haptic feedback have been used for robot teleoperation ~\cite{yashin2019aerovr}.

Teleoperation for mobile manipulators has been 
studied in~\cite{Santiago2019}, where control algorithms inspired by human interaction are developed so that the resulting teleoperation is intuitive and easy to use. In prior work on semi-autonomous mobile  manipulation~\cite{annem2019msec}, a mobile manipulator system called ADAMMS (Agile Dexterous Autonomous Mobile Manipulation System, see Fig. \ref{fig:system_hardware}) was developed. Limitations of this system are discussed in the next section.  


For evaluation purposes, many studies have relied on the time-to-completion (i.e., measuring the time elapsed between the operator starting and completing the task) measure as a qualitative measure for inferring system usability. Such times can also be systematically compared as a function of trial repetitions to estimate system learnability.  The assessment of mental effort using the NASA Task Load Index (NASA-TLX)~\cite{hart1988development} is a widely used tool to measure human performance and mental effort~\cite{annem2019msec}. In~\cite{steinfeld2006common}, several metrics of human-robot interaction (HRI) in response robot applications are identified that can be leveraged for comprehensive evaluations across a wide range of tasks and systems. In industry, standardized resources such as the International Organization of Standardization (ISO)  ISO 25010~\cite{ISO25010} are used to evaluate software systems and the operator’s response to the interface based on a number of qualitative metrics.

\begin{figure*}[h]
  \centering
  \includegraphics[width=1.8\columnwidth]{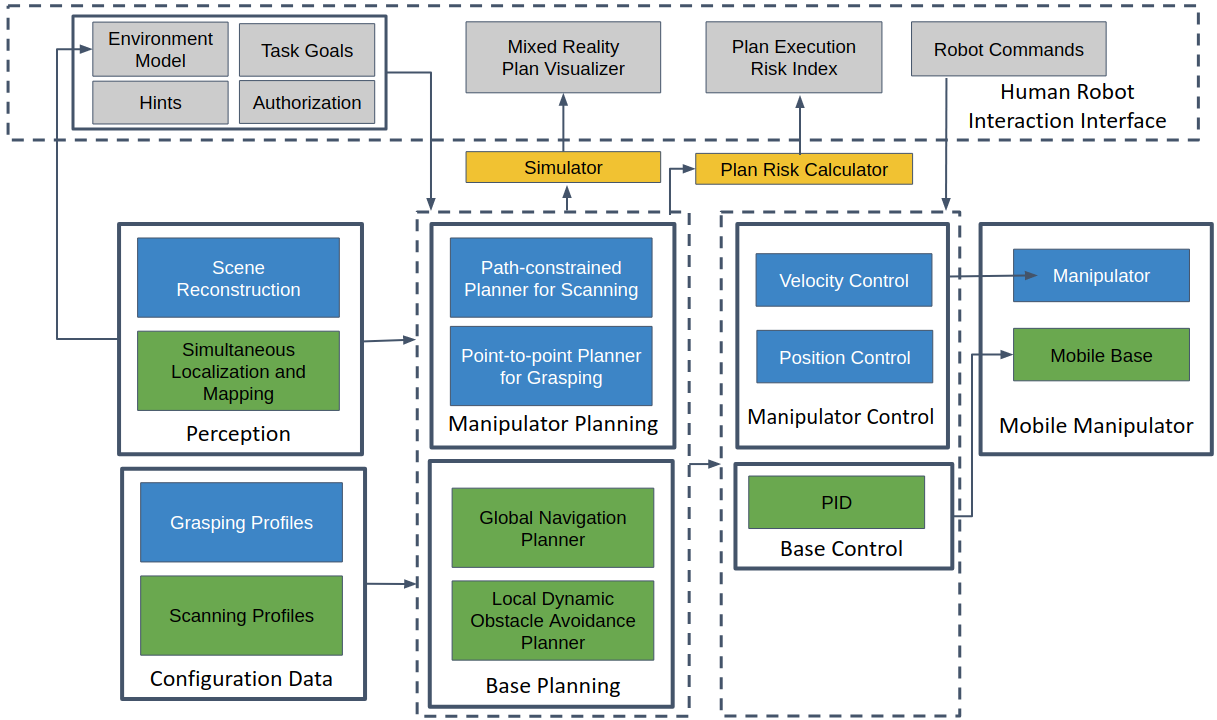}
  \caption{The architecture of the AI-HRI system for the mobile manipulator}
  \label{fig:sys_arch}
\end{figure*}

\section{System Requirements Development} 
Exploratory experiments were conducted with the ADAMMS 1.0 system to assess safety and efficiency issues with the first prototype~\cite{annem2019msec}. The operators were asked to complete machine tending tasks as efficiently as possible without compromising safety. The operators pointed out the following shortcomings after conducting these exploratory experiments:

\begin{enumerate}
\item 
A fixed depth camera mounted on the mobile base made it difficult for operators to quickly gain situational awareness for safely performing the machine tending task.
\item 
The discrete number of grasping directions made the grasping task time-consuming and non-intuitive. The absence of predictive collision avoidance forced the operator to move in a very cautious manner. This further slowed down progress on the manipulation task.       
\item 
Operators found it challenging to position the gripper at the desired pose using the goal configuration of the manipulator. This led to slow progress on the manipulation task. 
\item 
Operators found it difficult to select and specify waypoints for specifying the grasping motion for the manipulator. This resulted in increased time delays. 
\item 
Operators felt that the automatically generated mobile base plans were not safe because the resulting base motions were very close to the obstacles. Even small changes in the map resulted in collisions.  
\item 
The platform did not detect and avoid dynamic obstacles, i.e., people or chairs, which were not there in the recorded map. This led to safety concerns during run time.
\item 
The mobile base planner did not use turn-in-place moves and instead relied on arc motions to make turns.  This led to inefficient plans and task execution.   
\item 
Operators reported that it was difficult to verify the safety of generated motion plans for the mobile base. 
\end{enumerate}

The following requirements for ADAMMS 2.0 were developed based on the above observations: 

\begin{enumerate}
\item 
Use the arm to pan the depth camera to capture all the relevant features of the workspace to perform the manipulation task. In addition, incorporate a second color camera on the arm to show the gripper to the operator.   
\item 
Enable the operator to select a grasping direction of their choice and use predictive collision detection to report potential safety violations with the operator-selected grasping strategy.    
\item 
Enable the operator to position the gripper by selecting the gripper pose and automatically compute the manipulator configuration to realize the desired gripper pose.   
\item 
Enable the manual motion of the gripper based on live camera feed for grasping and check for collisions before implementing the manual motions.
\item 
Automatically generate mobile base plans that maintain safe distances from the obstacles.  
\item 
Detect dynamic obstacles and avoid collision with them. 
\item 
Incorporate in-place rotation actions during mobile base planning. 
\item 
Provide operators situational awareness to verify the safety of the mobile base motion plans.  
\end{enumerate}

This paper presents system architecture and design to address these requirements.   

\section{System Architecture}
To address the requirements identified in the previous section, the system was designed to find the best balance between reducing the task execution burden on human operators and ensuring operator safety in the presence of high uncertainty.  The architecture for the software system is illustrated in Fig.~\ref{fig:sys_arch}. The AI technology is used both for automating task execution (when possible) and providing information to humans to make risk-informed decisions to accept, refine, or abandon system-generated plans. The human operator has access to the environment model generated by the perception system. The human operator tasks the system by providing task goals. These tasks could be either to scan the environment or manipulate an object. The system automatically generates a plan to execute the given task.  The system uses a plan simulator to provide the operator with a mixed reality plan visualization. 

The system also computes the plan execution risk based on the uncertainty in the environment model. Uncertainty estimate are generated by performing multiple scans of the environment from many different camera poses. This data is fused together and spatial discrepancy in the fused data is used to estimate the uncertainty in the model built by the system. If the operator is satisfied with the plan, then the operator can authorize the controller to execute the plan. Conversely, if the operator is not satisfied with the plan, then the operator can offer hints to the planner by giving intermediate motion goals. The planner can generate new plans based on this additional information, and the plan's visualization and risks are presented to the operator. This process is continued until the operator is satisfied with the plan.  

If the operator believes that the system is unable to generate an appropriate plan, then they can abandon the system generated plan, and command the robot's motions using teleoperation. 

\section{Mobile Manipulator Platform Description} \label{sec:system}
The ADAMMS 2.0 system consists of a differential drive mobile robot, with a Universal Robots UR5 robotic manipulator physically mounted on the chassis. The manipulator is augmented with a Robotiq 2-fingered gripper attached to the tool flange. A multi-sensor suite is used to monitor the work volume, and consists of multiple depth cameras attached to both the mobile base and the manipulator, color cameras, a Lidar (light detection and ranging) area scanner, a 9 degrees-of-freedom inertial measurement unit (IMU), and encoders on the mobile base's wheels.  
\begin{figure}[h]
  \centering
  \includegraphics[width=1.0\columnwidth]{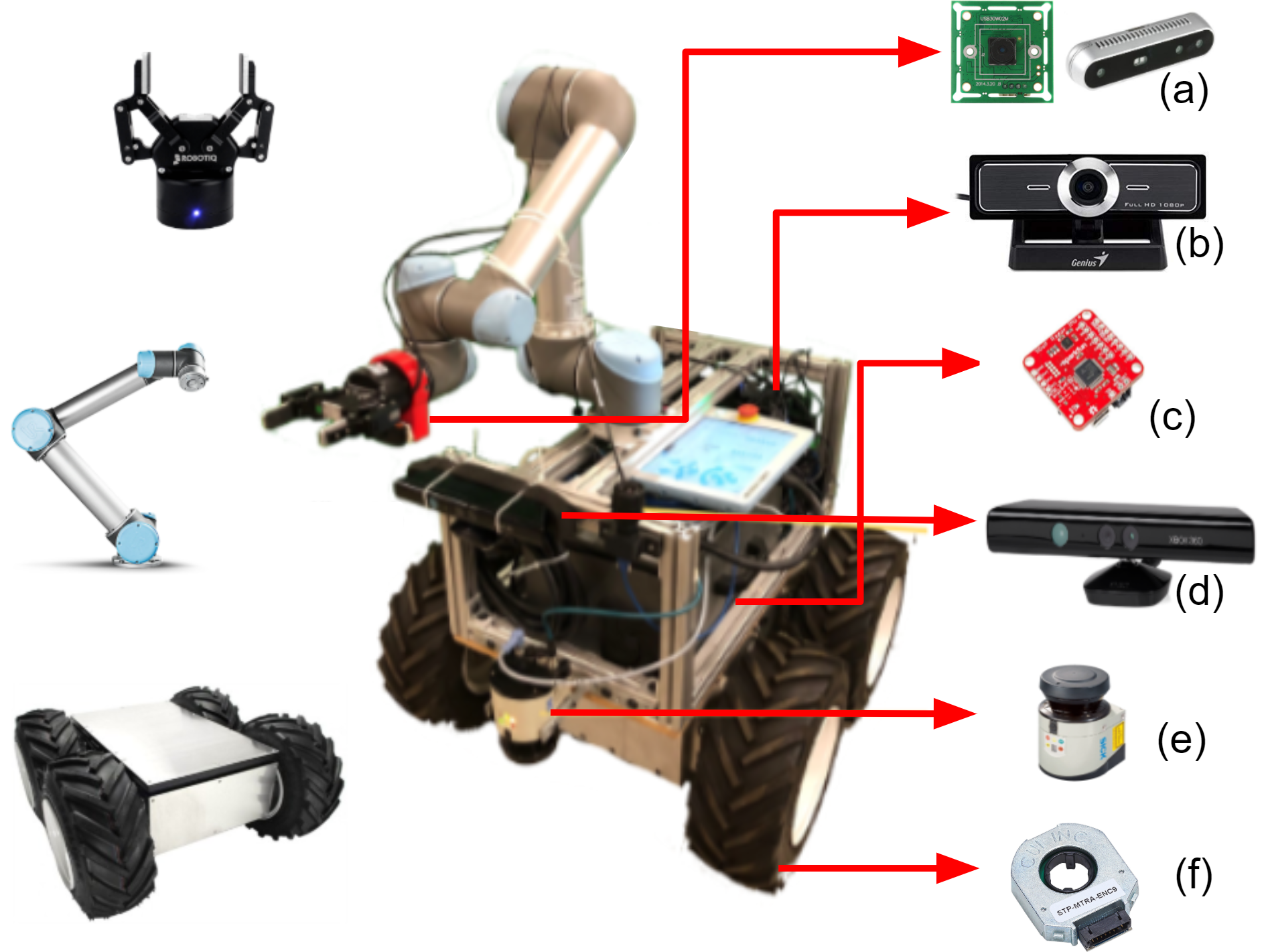}
  \caption{Hardware components of ADAMMS 2.0. (a) The gripper has been mounted with a camera and Intel Realsense camera (b) wide angle cameras mounted on the robot to understand the environment better (c) IMU for localization (d) kinect for mapping and localization (e) laser scanner for localization (f) wheel encoders for odometry}
  \label{fig:system_hardware2}
\end{figure}

\begin{figure}[!htbp]
  \centering
  \includegraphics[width=1.0\columnwidth]{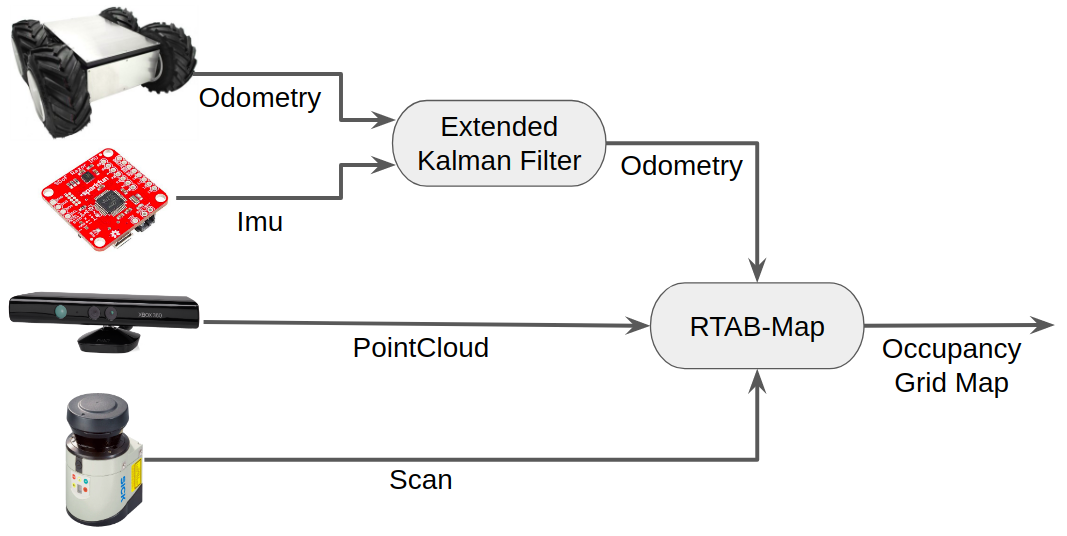}
  \caption{A simplified ROS message diagram for localization and mapping of mobile base}
  \label{fig:system_slam}
\end{figure}

As shown in Fig.~\ref{fig:system_hardware2}, the mobile manipulator system consists of three actuators, mobile base, manipulator, and gripper. The sensors are listed on the right. Compared to ADAMMS in~\cite{annem2019msec}, there are additional sensors for accurate localization of the mobile base with addition of an IMU and a lidar. These help in reducing the localization error and detecting dynamic obstacles in the local map.  For localization of the mobile base, RTAB-Map~\cite{labbe2019rtab} was used in ROS to take advantage of sensor data from the Kinect and 2D Lidar for mapping and localization. The Kinect has 3D depth data, but this data is subject to significant measurement uncertainty. In contrast, the 2D Lidar provides a more accurate and larger range of depth data. By fusing data from both sensors, both better mapping of the environment and robust feature detection can be achieved. Also, improved odometry estimates are provided as inputs to RTAB-Map by incorporating data from wheel encoders and an onboard IMU using an extended Kalman filter. The complete sensor setup for RTAB-Map is as shown in Fig.~\ref{fig:system_slam}. 

The camera (in green) at the end-effector in  Fig.~\ref{fig:system_hardware2}(a) is used to give the operator a view of the gripper fingers during grasping. 
The Intel Realsense camera also shown in Fig.~\ref{fig:system_hardware2}(a), on the end-effector is used to generate point cloud of the scene on demand. Using the depth camera for both the purposes is not desirable as having the gripper fingers in the view includes them in the point cloud. Also, it is necessary to have the gripper fingers in the view as it gives the operator an intuition towards manual grasping. 


\section{Performing Machine Tending Task with ADAMMS 2.0} \label{sec:gui}
\begin{figure*}
  \centering
  \includegraphics[width=1.8\columnwidth]{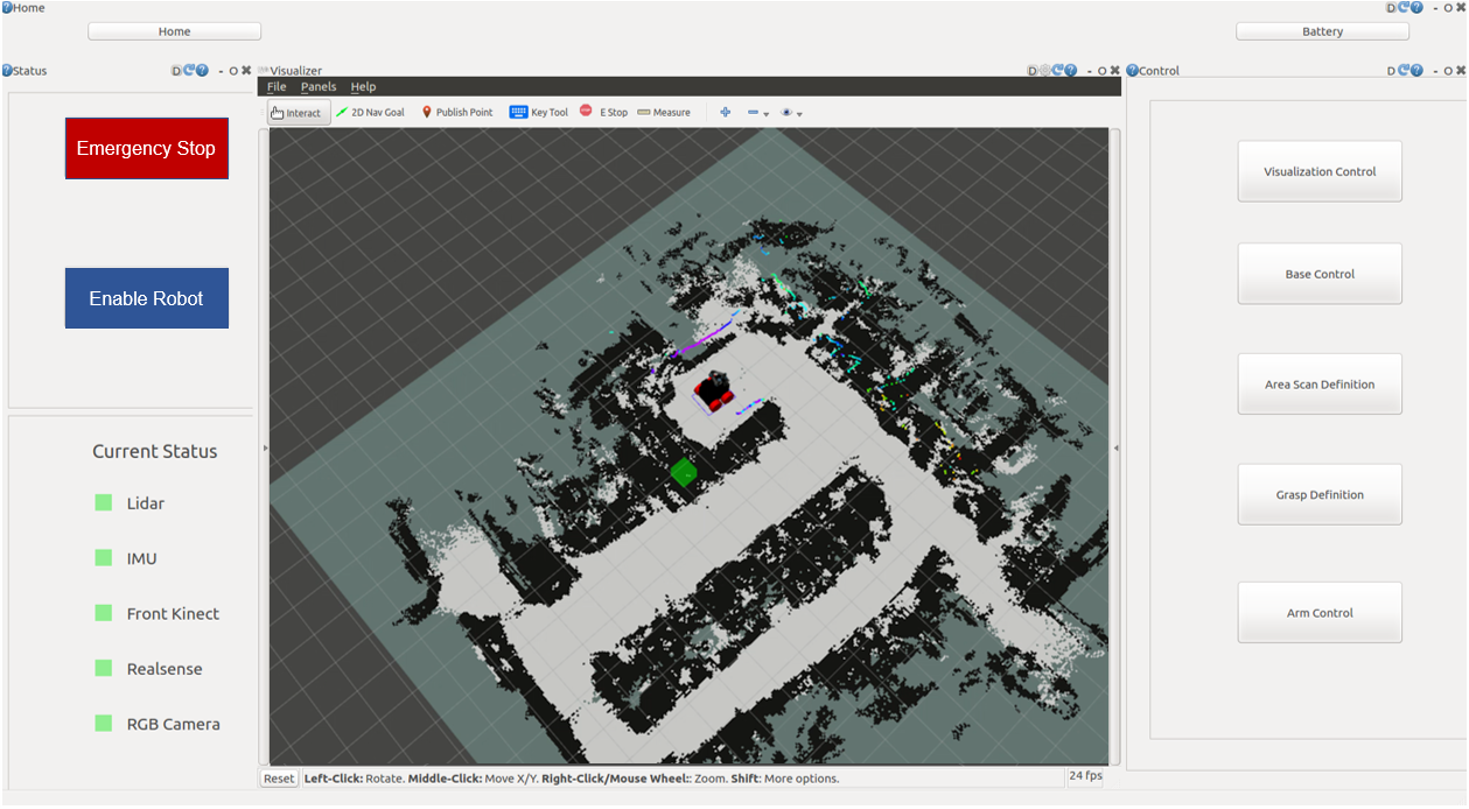}
  \caption{The home screen of the graphical operator interface for machine tending tasks using ADAMMS 2.0. The left panel has the common control used for enabling and stopping the robot, as well as a status panel showing the state of all the sensors on-board. The right side of the panel shows different control screens which user can access.}
  \label{fig:gui_main}
\end{figure*}

In this section, the different features that have been introduced are discussed in the context of supporting machine tending tasks.
\Dgreen{Fig. \ref{fig:gui_main} shows the home page of the graphical interface used by the operator. The left side of the home screen includes major functionalities like enabling the robot and emergency stop buttons, and status of different sensors on the robot. The right side is used to select a control mode  from the five different options. Base control and arm control modes provide the control panel for the motion of the base and the arm, respectively. The Arm Scan definitions option, in the panel is used to generate scanning profiles offline. These profiles are later selected by the operator while tasking robot to perform scanning operation of the scene (see Fig. \ref{fig:gui_scanning_profiles}). Similar to scanning, grasp definitions allows the user to generate various grasp configuration which can later be used online. Finally, the visualization control mode allows the operator to switch between different sensor feeds while operating the platform.
}
We will use a  mock-up of a 3D printer to illustrate machine tending task 
\Dgreen{by ADAMMS 2.0 using this interface. }

\subsection{Mobile Base Motion Towards the Machine}\label{subsec:mb}

The mobile base moves from the starting location to an operator set goal location as the desired position and orientation of the mobile base. A GUI for HRI is used with the AI aspect of it being the motion planning for the mobile base. A mobile robot motion planning package in ROS called \textit{move\_base} is used for planning. Here, the time-elastic bands (TEB)~\cite{rosmann2017kinodynamic} planner is used for local planning. In TEB, the optimal trajectory is efficiently obtained by solving a sparse scalarized multi-objective optimization problem. We adjust the weights of different objectives to produce the desired behavior in case of conflicting objectives. Since local planners such as TEB often get stuck in a locally optimal trajectory depending on the initial seed, a subset of admissible trajectories in different homotopy classes is optimized in parallel. The local planner is able to switch to the current globally optimal trajectory among the candidate set. For local planning, it uses a real-time map generated using the lidar and plans the path in real-time. The GUI also provides a stop button that the operator can utilize to manually move the robot once the robot comes close to the goal. Further, the manual motion also helps in fine-tuning the pose of the mobile robot near the 3D printer. The TEB planner used as compared to the Dynamic-Window Approach used in~\cite{annem2019msec} provides a much stable trajectory with multiple parameters to be set by the operator. 

\subsection{Manipulator Motions for Grasping}
\label{subsec:mmlabel}
\begin{figure}[h]
  \centering
  \includegraphics[width=1.0\columnwidth]{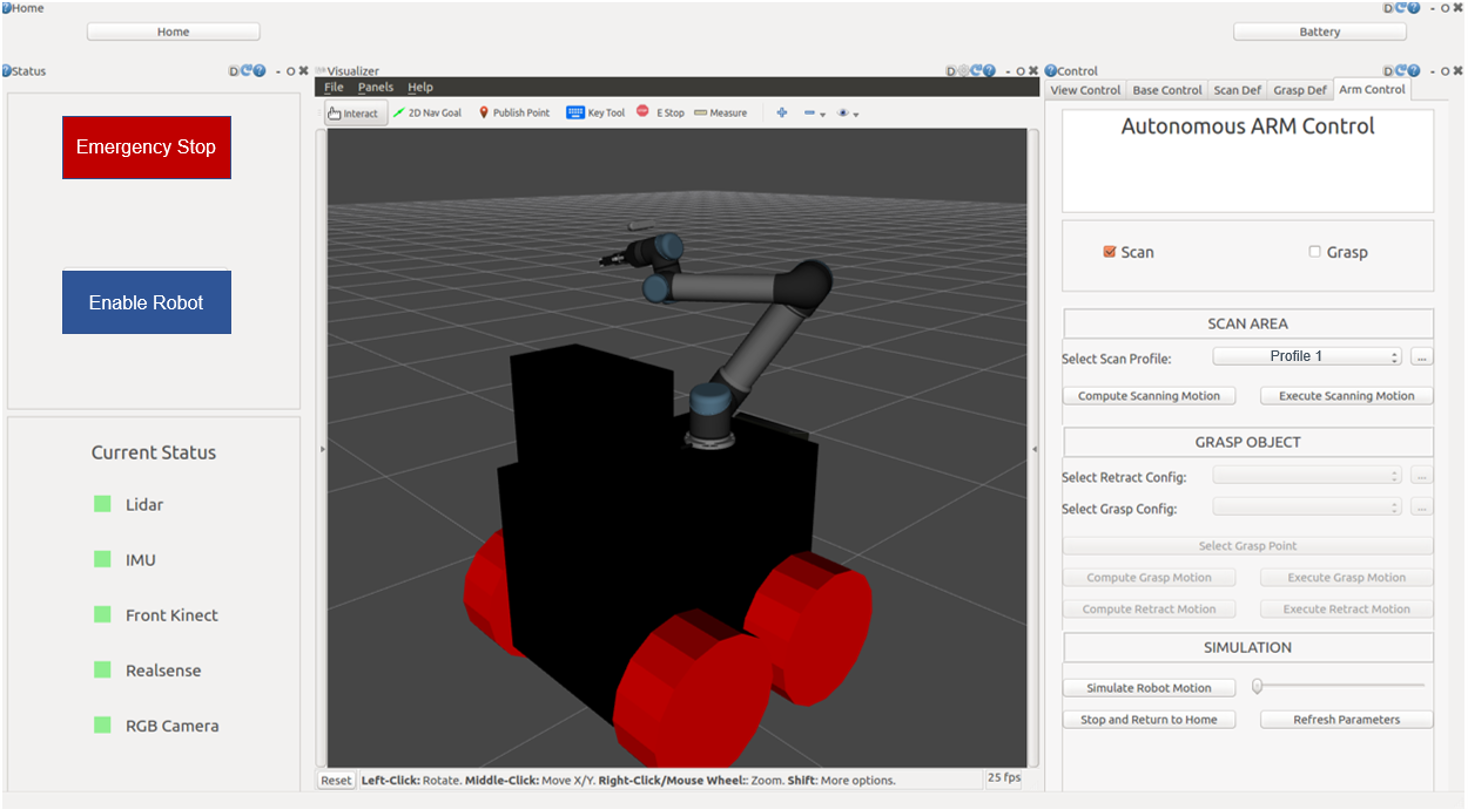}
  \caption{The interface showing the Automated Arm Control Panel used for scanning and grasping operation. The user can enable one of the modes at a time, and select a pre-defeined scanning, grasping, retract profile to rapidly provide high-level input to low level planning algorithms to generate robot motions.}
  \label{fig:gui_arm_control}
\end{figure}

After the mobile base reaches an operator-defined location near the machine, the manipulator initiates motion for grasping the part. \Dgreen{The Automated Arm Control Panel of the interface is used for scanning and grasping operation, as depicted in Fig. \ref{fig:gui_arm_control}. }

\begin{figure}[h]
  \centering
  \includegraphics[width=1.0\columnwidth]{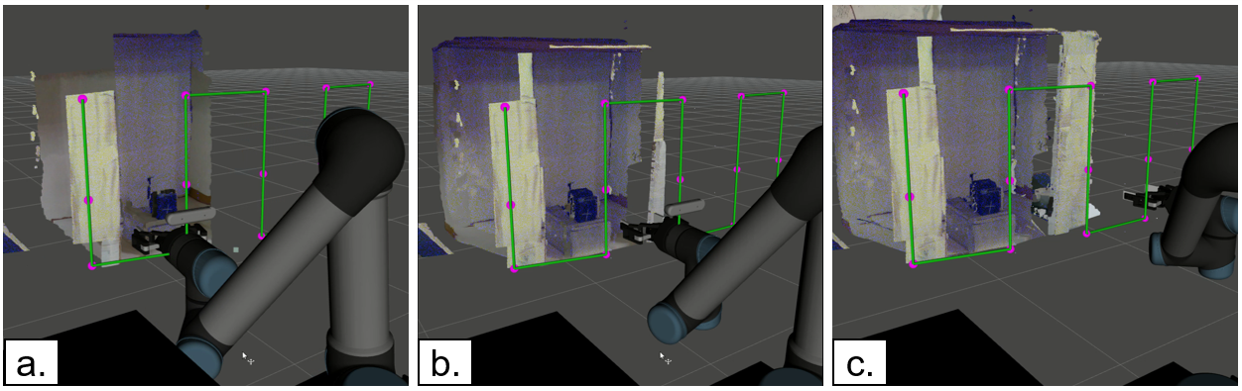}
  \caption{An of scene scanning using automatically generated robot trajectory for a selected scanning profile. The robot takes pointcloud data at each waypoint (magenta color) on the path (green color) and filters and stitches it to generate the scene model. }
  \label{fig:gui_stitching}
\end{figure}

\begin{figure}[h]
  \centering
  \includegraphics[width=1.0\columnwidth]{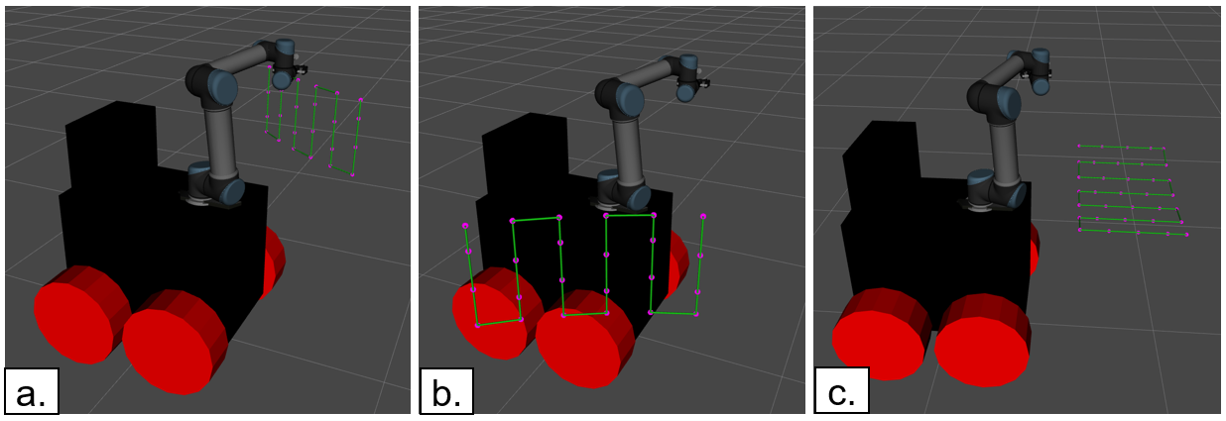}
  \caption{The figure shows few types of pre-defined scanning profiles that can be used by the user to scan the scene. These scanning profile are generated beforehand, and the operator just need to select a profile during operation of the robot. The current architecture auto generates the camera-path (shown in green) using a developed planner and any mesh profile. This eliminates the need of teaching robot scanning positions.}
  \label{fig:gui_scanning_profiles}
\end{figure}

The first step is to capture a point cloud of the 3D printer and the surroundings. An Intel Realsense mounted on the gripper is used to capture a point cloud of the environment to produce a 3D model of the scene. The gripper movement is planned to capture multiple point clouds and stitch them together to reconstruct the scene, \Dgreen {as shown in Fig.~\ref{fig:gui_stitching}. The interface provides few pre-defined scanning profiles for this scanning process; some examples are shown in Fig. \ref{fig:gui_scanning_profiles}. The operator can readily choose from these options. Once a profile is selected, the system auto-generates the manipulator's trajectory using a path-constrained motion planner \cite{ICRA_2019_Kabir} such that the camera moves to go through the waypoints, capture and stitch the pointclouds. Each profile is generated using a mesh model which allows the user the flexibility to import any shape of mesh and generate a scanning profile.}

The quality of the reconstructed scene by stitching point clouds from Realsense camera is important since it is used to perform grasping of objects.
We perform multiple measurements of the stitched point cloud by using different sets of waypoints of the manipulator, and we analyze them to generate an uncertainty estimate of the 3D reconstructed scene. This way, we construct a single point cloud of the scene, where each point has an expected location, and a standard deviation error along the estimated surface normal direction. It gives humans insight on the level of uncertainty in different regions of the scene. If the uncertainty is above a certain threshold around the object of interest, the system can alert the human operator to use caution.

\begin{figure}[!htbp]
  \centering
  \includegraphics[width=1.0\columnwidth]{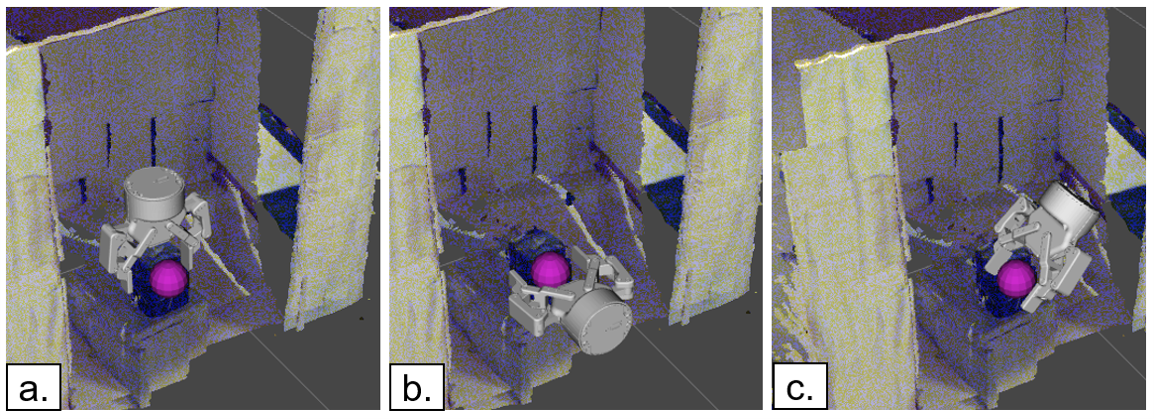}
  \caption{An example of setting the pre-grasp pose. The grasping pose is selected from the offline generated grasping profile repository and altered for grasping a particular object online. This allows the user to rapidly operate the robot, and not use gripper jogging controls to decide grasp configs for grasping each object.}
  \label{fig:gui_grasp_configs}
\end{figure}

Once the model of the machine and the nearby environment is reconstructed with reasonable quality, the next task is to grasp the part. For this purpose, the operator can set a pre-grasp pose for the gripper. \Dgreen{First, the operator selects a grasp position in the pointcloud after identifying the object to grasp. For rapid operation, a suitable grasping pose can be selected from a list of pre-generated grasp profiles, as shown in Fig. \ref{fig:gui_grasp_configs}. After that, the operator can further fine-tune the gripper configuration by using the jogging functionality, and finalise the grasp pose.}


\begin{figure}[h]
  \centering
  \includegraphics[width=1.0\columnwidth]{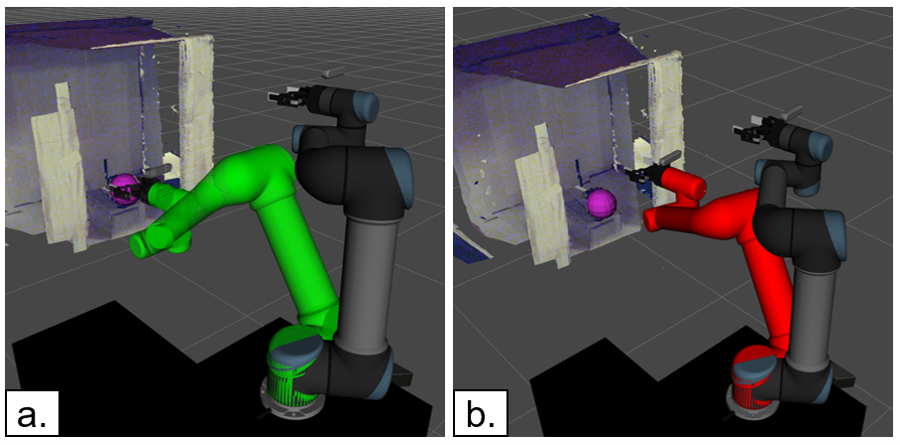}
  \caption{Visualization of the selected IK solution and it's collision checking result}
  \label{fig:gui_ik_collisioncheck}
\end{figure}

After the pre-grasp pose of the gripper is chosen, a finite number of distinct inverse kinematic (IK) solutions are generated for the manipulator. Solutions are sorted based on the distance from the current configuration so that the first solution is usually the best solution. However, the operator can choose a solution other than the first one to give more flexibility. Each of these solutions can be visualized for collisions. As shown in Fig. \ref{fig:gui_ik_collisioncheck}, if the solution is neither in collision with the robot itself nor the environment, the robot is visualized in green (left). Otherwise, the robot turns red (right). A pre-grasp configuration of the manipulator once set, is achieved using motion planning. Motion planning of a high degree of freedom systems like manipulators is challenging due to high dimensional state space in which a search has to be performed to reach from the starting configuration to the goal configuration. Based on the point cloud data of the system's surroundings, planning for such motion is one application for which AI is leveraged to assist the human operator. \Dgreen{For such motion planning queries, we have used a point-to-point planner \cite{CASE_2018_Kabir,IROS_2019_Rajendran} which generates a smooth robot trajectory from the current robot configuration to grasp configuration. This particular planner was developed to generate deterministic low-cost robot trajectory in confined work spaces which suits well for machine tending operations.} The operator can observe the planned trajectory before executing it with the real robot. The operator can choose to execute the trajectory or plan another one. 

After reaching the pre-grasp configuration, the operator can manually move the gripper in its coordinate frame to fine-tune and reach the part. Visual feedback is provided by the 3D rotatable UI and the camera at the end-effector.
The GUI has buttons to move linearly forward, backward, right, left, up, or down in the tool frame. Since the motions to fine-tune are small, such end-effector motions are feasible. Moreover, once a button for any direction is clicked, the motion is verified to be collision-free and then executed. This ensures the safety of the robot and the 3D printer. 
After the grasping is done, the manipulator needs to be bought to the initial configuration so that the mobile base motion can be resumed. For this, use OMPL for motion planning as before. However, care must be taken to ensure that the part does not collide with the other objects in the workspace when the manipulator is in motion. For this, a collision sphere is used, 
which engulfs the entire part.


\subsection{Mobile Base Motion Towards the Goal}
Once the grasping has been done, and the manipulator has retracted towards its initial position, the mobile base can move towards the goal location, where the part needs to be transported. The operator provides the same commands as discussed previously. 
Similarly, the drop functionality is achieved using the aforementioned tools for moving the manipulator and opening the gripper. 

\section{Experiments and Results} \label{sec:result}

\subsection{Comparison with ADAMMS 1.0}
The first modification is mounting a depth camera on the end-effector of the manipulator and capturing a point cloud from operator set angles. This results in a denser point cloud, compared to using a fixed depth camera on the mobile base. 


The next modifications required the operator to have flexibility in 
\Dgreen{determining how the gripper will approach for grasping an object. The operator can select and adjust the pre-grasping pose of the gripper, and a set of inverse kinematic solutions is provide for the operator-set gripper pose.
}
Here, multiple types of objects can be grasped with the corresponding collision-free inverse kinematic solutions.


The manipulator is moved manually for grasping after reaching the operator set pre-grasping configuration. 
However, making sure that these manual motions are collision-free before executing them is critical \Dgreen{since it can be difficult for the operator to perceive from the camera views and the interface. If the manual jogging motion may result in a collision, the system ceases the motion beforehand, and gives collision warning to the operator.}

\begin{figure}[h]
  \centering
  \includegraphics[width=1.0\columnwidth]{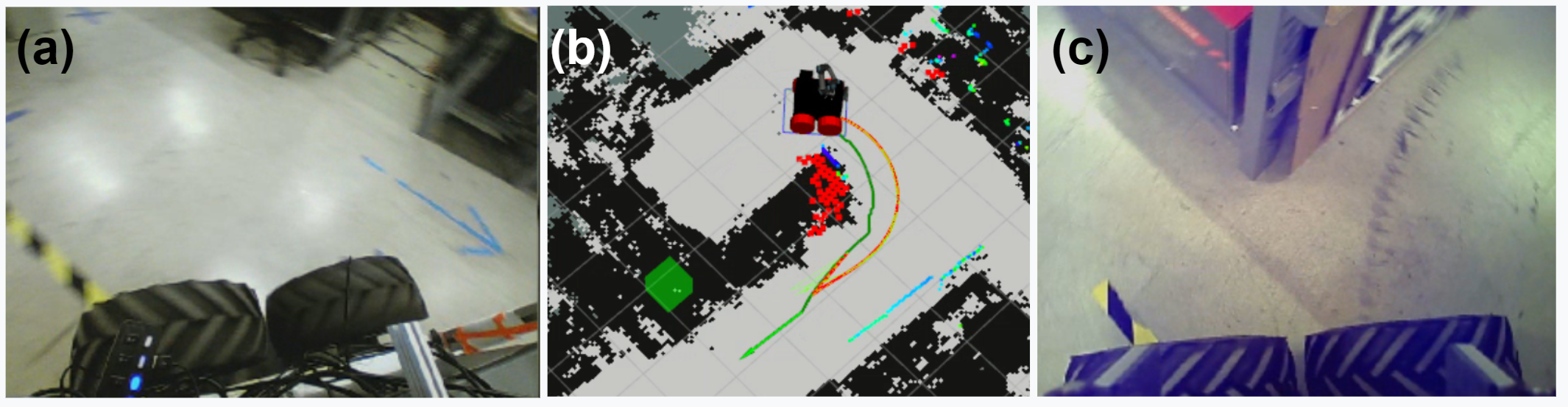}
  \caption{(a) The wide angle camera viewing the left (c) and right of the mobile base respectively. (b) The mobile base local path in red. It can be seen that this path is near the center of the passage.}
  \label{fig:mb_safe}
\end{figure}

The next requirement was to make sure that the mobile base avoids collisions due to small changes in the map. Once the environment is mapped, this map can be used so long as there are no significant changes to the environment; however, small displacements of tables, chairs need to be taken into account. The TEB planner is based on optimization where the center of a passage for the path is a local minimum. It generates paths, as shown in Fig.~\ref{fig:mb_safe}(b) in red, which passes near the center of the passages. Moreover, two wide-angle cameras can be used to monitor any obstacle. Furthermore, the operator has an option to stop the motion generated by the planner and use these wide-angle cameras to find the way around obstacles.
The wide-angle cameras also provide the operator the required  situational awareness.
    
For avoiding dynamic obstacles, in~\cite{annem2019msec}, a Lidar would stop the motion of the mobile base if a previously-unmapped obstacle was detected in its way. The operator would have to manually move the robot, or provide a goal point again after the dynamic obstacle had moved. In this work, a local TEB planner is used, which uses the Lidar information to modify the existing path after detecting a dynamic obstacle. This greatly reduces the operation time and the burden on the operator to move the mobile base. The path in red in Fig.~\ref{fig:mb_safe}(b) is one such local path.

\begin{figure}[h]
  \centering
  \includegraphics[width=1.0\columnwidth]{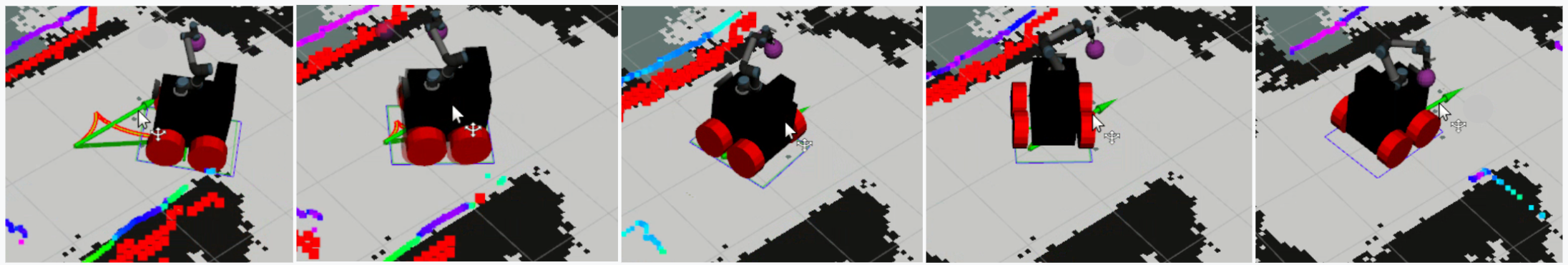}
  \caption{The mobile base rotating in place.}
  \label{fig:mb_rotate}
\end{figure}  

The final requirement was that the mobile base planner should allow for rotation in place, which is necessary, especially in narrow passages (Fig. \ref{fig:mb_rotate}). This is possible due to the parameters in TEB, which can be set for the minimum radius of turning. This greatly enhances the motion capabilities of the platform as compared to in~\cite{annem2019msec}.

\subsection{Initial User Trials}
First-level trials were conducted with 6 volunteers taken from the pool of people working at our lab. These volunteers all have some experience with robotic systems. Amongst these volunteers, 2 were female graduate students (mean age $\sim$30) with an engineering background, but limited robotics experience. The remaining 4 participants were male graduate students in engineering with 2 (mean age $\sim$25) having significant experience with robots, and 2 (mean age $ \sim$27) with limited robotics experience. 

A metric was defined to measure the ease of use of the system based on the time taken by a new operator to complete the task as compared to an expert operator. This metric is inversely proportional to the difference between the time taken by a trial operator and the expert operator to complete the same task. Here, for a particular task since the time taken by the robot to move is similar for the expert and trial operators, the time difference measures the effort a new operator has to put in to understand the GUI and operate the system.

All the operators attempt the same task of guiding the robot to a 3D printer placed in the lab and retrieving a part. The expert operator took 14 minutes to complete this particular task. Without any prior training, the average time taken by the other operators was 24 minutes. In the second attempt, after having limited experience with operating the system, the operators took on average 18 minutes to complete the task. This shows that the usability of the system significantly increases with the experience of operating the system. The main reason for this change was gaining a better understanding of the GUI and robot capabilities. 

There were two important decision-making factors that were observed to have significantly reduced the task completion time. The first factor was having a better insight into where to place the mobile base so that manipulator can reach to grasp the part. For a new operator, without an intuition of the reachability of the mobile manipulator, it took longer to complete the task as compared to an experienced operator. The second factor was that for a new operator, understanding whether the part is within the gripper fingers is time-consuming as he/she has to go back into the 3D view to observe the gripper location after each manual motion. With experience, the operator could precisely grasp the object by manual motion. This rudimentary operator trials resulted in an overall positive feedback for changes made to the system. 

\section{Conclusions \& Future Work} \label{sec:conclusion}
This paper presents the ADAMMS 2.0 mobile manipulation system for performing machine tending tasks in a safe and efficient manner. In the prior work, a preliminary version of this system was presented with operator trials, and several requirements were compiled based on the feedback received. In this work, all of these requirements are addressed. The use of AI enables automated motion planning from the high-level task description. The AI is also leveraged to simulate plans and compute risks. These plans are presented to the human operator. This enables humans to intervene when collision risk is high due to the uncertainty in the environment model. The system design allows the human operator to simply authorize the execution of automatically generated plans when risk is low and perform teleoperation when the risk is high. This design allows humans to archive operational efficiency without compromising safety.     

Extensive operator trials are planned to gauge the ease of use of the system. Scheduled wide-scale human trials for this work were postponed due to the global COVID-19 pandemic. Based on the feedback from the limited operator trials, the current plan integrates reachability maps to generate areas where the mobile base should be located so that the part is reachable for the manipulator, and the operator makes quick decisions. Moreover, the work will be expanded to handle different types of machines. Machines with doors require constrained manipulator motions as well as placement of the mobile base so that such motion is possible. Furthermore, there is also a need to have special mounts on the gripper such that the door of the machines can be opened easily. 

{\bf Acknowledgements:}
This work was supported in part by National Institute of Standards and Technology Award $\#$ 70NANB15H250N. Opinions expressed are those of the authors and do not necessarily reflect opinions 
of the sponsors.


{\bf Disclaimer:} Commercial equipment and materials are identified to adequately specify certain procedures. In no case such identification implies recommendation or endorsement by University of Southern California or the National Institute of Standards and Technology, nor does it imply that the materials or equipment identified are necessarily the best available for the purpose. 



\bibliographystyle{aaai}

\bibliography{main}

\end{document}